\documentclass{article}

     \PassOptionsToPackage{numbers, compress}{natbib}

\pdfoutput=1

     \usepackage[preprint]{neurips_2022}

\usepackage[utf8]{inputenc} 
\usepackage[T1]{fontenc}    
\usepackage{hyperref}       
\usepackage{url}            
\usepackage{booktabs}       
\usepackage{amsfonts}       
\usepackage{nicefrac}       
\usepackage{microtype}      
\usepackage{xcolor}         
\usepackage{graphicx}

\title{Discovering Dynamic Functional Brain Networks via Spatial and Channel-wise Attention}

\author{%
	Yiheng Liu \\
	School of Physics \& Information Technology\\
	Shaanxi Normal University\\
	Shaanxi Normal University, Xi'an, China 710119 \\
	\texttt{liuyiheng@snnu.edu.cn} \\
	 \And
	 Enjie Ge \\
	 School of Physics \& Information Technology \\
	 Shaanxi Normal University \\
	 \texttt{ejge@snnu.edu.cn} \\
	 \And
	 Mengshen He\\
	 School of Physics \& Information Technology \\
	 Shaanxi Normal University \\
	 \texttt{hems@snnu.edu.cn} \\
	 \And
	 Zhengliang Liu\\
	 University of Georgia\\
	 \texttt{zl18864@uga.edu}\\
	 \And
	 Shijie Zhao\\
	 Northwestern Polytechnical University\\
	 \texttt{shijiezhao@nwpu.edu.cn}\\
	 \And
	 Xintao Hu\\
	 Northwestern Polytechnical University\\
	 \texttt{xintao.hu@gmail.com}\\
	 \And
	 Dajiang Zhu\\
	 University of Texas at Arlington\\
	 \texttt{dajiang.zhu@uta.edu}\\
	 \And
	 Tianming Liu\\
	 University of Georgia\\
	 \texttt{tianming.liu@gmail.com}
	 \And
	 Bao Ge \\
	 School of Physics \& Information Technology \\
	 Shaanxi Normal University \\
	 \texttt{bob\_ge@snnu.edu.cn} \\
}

\begin{document}

	\maketitle

\begin{abstract}
Using deep learning models to recognize functional brain networks (FBNs) in functional magnetic resonance imaging (fMRI) has been attracting increasing interest recently. However, most existing work focuses on detecting static FBNs from entire fMRI signals, such as correlation-based functional connectivity. Sliding-window is a widely used strategy to capture the dynamics of FBNs, but it is still limited in representing intrinsic functional interactive dynamics at each time step. And the number of FBNs usually need to be set manually. More over, due to the complexity of dynamic interactions in brain, traditional linear and shallow models are insufficient in identifying complex and spatially overlapped FBNs across each time step. In this paper, we propose a novel Spatial and Channel-wise Attention Autoencoder (SCAAE) for discovering FBNs dynamically. The core idea of SCAAE is to apply attention mechanism to FBNs construction. Specifically, we designed two attention modules: 1) spatial-wise attention (SA) module to discover FBNs in the spatial domain and 2) a channel-wise attention (CA) module to weigh the channels for selecting the FBNs automatically. We evaluated our approach on ADHD200 dataset and our results indicate that the proposed SCAAE method can effectively recover the dynamic changes of the FBNs at each fMRI time step, without using sliding windows. More importantly, our proposed hybrid attention modules (SA and CA) do not enforce assumptions of linearity and independence as previous methods, and thus provide a novel approach to better understanding dynamic functional brain networks.
\end{abstract}

\section{Introduction}
Brain activity is measured by capturing signals of neural activities. Functional magnetic resonance imaging (fMRI) \cite{logothetis2008we} is a non-invasive imaging method with no side effects that measures the blood-oxygen level dependence (BOLD). It is a common approach used to indirectly observe neural activities of the brain over time \cite{2}. FMRI is a very useful source of information for understanding the brain's intrinsic functional network structure and human brain functions \cite{3,4}. Functional brain networks (FBNs) can be defined as collections of regions displaying functional connectivities  \cite{archbold2009neural,4,lv2017task}. Because these FBNs will change over time  \cite{allen2014tracking,ma2014dynamic,hutchison2013dynamic,leonardi2015spurious}, analyzing the patterns of functional networks over time is critical to understand how the brain works. 

Most studies extract FBNs from entire fMRI signals, assuming that FBNs remain constant throughout the fMRI scanning process. However, the FBNs are in fact dynamic, fading in and out over time, because the brain is naturally dynamic, characterized by continuous discharge activities at the neuron level. The existing methods mainly adopt a sliding window scheme to detect time varying FBNs \cite{ma2014dynamic,DFBNSDL,4,yuan2018spatio,calhoun2016time,kiviniemi2011sliding,leonardi2015spurious}. They usually carefully set a window size and a sliding step in order to detect time varying patterns, but FBNs exhibit transient states of brain activity\cite{44}, which makes it difficult for these methods to identify the transient FBNs with sliding windows. Furthermore, FBNs can't be selected automatically in previous methods, and the number of FBNs needs to be set manually in a heuristic or experiential manner. Another limitation is the assumption of linearity or independence in most of existing methods \cite{5,6,10}. For example, independent component analysis (ICA) \cite{6,7,9} regards the identification of functional brain networks as a problem of blind source separation, that is, it is assumed that the observed signals are composed of the linear superposition of independent blind source signals. Spatial ICA \cite{8} assumes that each blind source component is spatially independent. The generalized Linear model (GLM) \cite{5} assumes that the observed signal is a linear superposition of the task design signal, and considers each voxel as an independent variable. Recently, some deep learning methods can learn the better nonlinear features of fMRI signals, but they still recover FBNs with linear LASSO regression \cite{14,15,16,17,18,19,20,21,26,30}. Overall, these methods oversimplify the relationship between BOLD signal changes and the heterogeneity of neuronal activity within voxels, and cannot reflect the complex functional interactions among brain regions.  

To overcome these problems, we propose a novel method to discover dynamic FBNs, which is named spatial and channel-wise attention autoencoder (SCAAE), inspired by the visual attention mechanisms\cite{SENet,CBAM} in the field of image recognition.  The method is based on the fact that fMRI signal in a neuron or brain region is not only induced by its internal neural activities, but also shaped by the interactions with other neurons or brain regions \cite{huang2016latent}. Therefore, it is necessary to increase the receptive field of the model. We use 3D convolution neural networks with large kernel size to encode and reconstruct the fMRI signals, spatial-wise attention (SA) to directly detect the FBNs in the whole brain without linear and independent assumptions, and channel-wise attention (CA) to weigh the important channels to select FBNs automatically. Because neural activities lead to differences in fMRI signals at each time step, the attention mechanism will focus on regions of neural activities.

In short, our contributions are summarized as follows:

1. The proposed SCAAE method can discover dynamic transient FBNs at each time step without using sliding windows, thus providing a new perspective for functional network analysis.

2. SCAAE can directly recover spatially interactive FBNs via a spatial-wise attention (SA) module, thus being free from linear and independent assumptions, and can select FBNs automatically via a channel-wise attention (CA) module, thus avoiding manual interventions.

3. The decoder and partial encoder components are not needed in forward inference. SCAAE only needs the output of the attention modules. This process of constructing the FBNs is much faster than previous work because it dispenses with learning the coefficient matrix iteratively.

\section{Related Works}
	\subsection{Functional Brain Network}
	At present, the commonly used methods to construct functional networks based on fMRI data include: general linear model (GLM) \cite{5}, independent component analysis (ICA) \cite{6,7,8,9} and sparse dictionary learning (SDL) \cite{10,11,12,13}. Because these methods require less data and are more stable than deep learning methods, they are widely used in clinical practices. Among them, GLM is used to model task-based fMRI (tfMRI), ICA is used to model resting state fMRI (rsfMRI), and SDL can model both tfMRI and rsfMRI. To discover dynamic FBNs, some works introduced sliding windows \cite{ma2014dynamic,DFBNSDL,allen2014tracking,yuan2018spatio,calhoun2016time,kiviniemi2011sliding,leonardi2015spurious}. However, limited by the shallow nature of these traditional machine learning methods and the sliding windows, they cannot extract fMRI intrinsic features and are unable to discover FBNs at each time step. In addition, These methods have their own assumptions about the fMRI data. For example, GLM relies on task design, ICA assumes an independent relationship between the different components, and SDL relies on sparsity assumption. 
	
	\subsection{Autoencoder}
	
	Deep learning has powerful representation capabilities, therefore it is often used to extract high-level features from low-level raw data. The output of each layer in deep learning model can be viewed as a new representation of the raw data and it has been proven in multiple tasks that deep learning methods are superior at learning high-level and mid-level features from low-level raw data.  \cite{bengio2013representation,lecun2015deep}. Due to the lack of labeled data, many methods based on deep learning use autoencoders (AE) to extract high-level features. Some examples include Deep Belief Nets (DBN) \cite{14}, Convolutional Autoencoder (CAE) \cite{15,16,17,26}, Recurrent Autoencoder (RAE) \cite{18,19}, Transformer Attention Autoencoder \cite{20} and Variational Autoencoder (VAE) \cite{21,30}. Although deep models can extract better feature representations than traditional methods, constructing FBNs with lasso regression does not stay away from the constraints of linearity assumption. 
	
	Previous work using autoencoders to construct FBNs is based on correlation. Prior methods use the encoders to extract the global features $Z$ of the raw fMRI signals $X$:
    \begin{equation}
    	Z_{T \times K} = encoder(X_{T \times M})
    \end{equation}
    \begin{equation}
    	min||X - ZW_{K \times M}||^{2}_{2} + \lambda||W||_{1}
    \end{equation}
    
    Where $T$ refers to the time series length of the fMRI signals, $K$ refers to the dimension of the extracted features, $M$ refers to the number of voxels of the raw fMRI signals. The extracted features $Z$ are treated as source components of the raw fMRI signals. $Z$ is used to predict the raw fMRI signals. $W$ refers to the coefficient matrix of the lasso regression. The matrix has $K$ rows, and the coefficients of each row represent the linear correlation of the source components $Z$ and the corresponding fMRI voxels. For voxels that are uncorrelated with the components, their coefficients are set to 0 by L1 regularization. Finally, each row of the coefficient matrix is inversely transformed to the 3D brain image space. This is done with masking in data preprocessing \cite{38}. Thus, FBNs has a neuroanatomically meaningful context.
	
	\subsection{Attention Mechanisms}
	Attention Mechanisms have been applied to multiple tasks such as machine translation \cite{gehring2017convolutional,vaswani2017attention}, image classification \cite{SENet,CBAM}, object detection \cite{dai2017deformable}, semantic segmentation \cite{fu2019dual}, medical image processing \cite{unet2018attention} and other tasks  with impressive performance.
	
	The voxels of fMRI images can capture the intensity of neural activities. Since attention mechanisms have great ability to select features, capturing brain functional activities with attention mechanisms is very suitable. We adopt a spatial-wise attention (SA) module to capture the FBNs in the whole brain and a channel-wise attention (CA) module to weigh the importance of the FBNs for selecting FBNs automatically.

\section{Methods}
\label{2}
\begin{figure}[h]
	\centering
	\includegraphics[width=\textwidth]{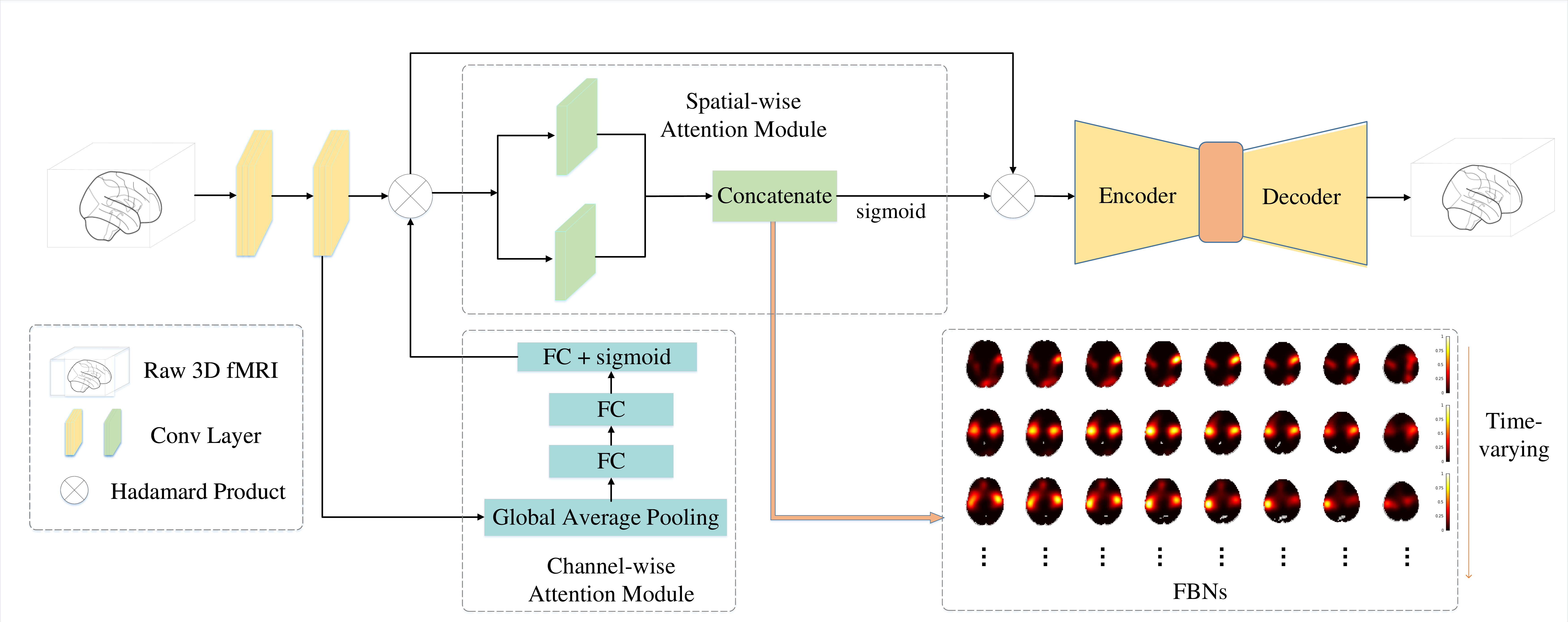}
	\caption{The key contribution of SCAAE is to construct FBNs via attention mechanism. A Channel-wise Attention (CA) module is used to weight the importance of FBNs for selecting FBNs automatically and a Spatial-wise attention (SA) module is used to capture the neural activities regions in the whole brain.}
	\label{fig1}
\end{figure}

The goal of our method is to construct FBNs via attention mechanism. The whole framework is shown in Figure~\ref{fig1}. We use the 3D convolutional neural network to encode and decode the raw fMRI signals. In our experiments, the network structure of down-sampling encoding and up-sampling decoding to reconstruct the raw fMRI signals is not critical, but inserting the attention modules into the specific layers can significantly affect the construction of FBNs. We adopt a spatial-wise attention module and a channel-wise attention module in the encoder part for discovering the functional brain networks. In section \ref{2.1}, we will explain why the FBNs can be discovered by the attention mechanism. In section \ref{2.2}, we will illustrate the details of constructing FBNs via attention mechanism. Our experimental results and details will be shown in the section \ref{3}.

\subsection{Attention}
\label{2.1}
The spatial attention is promising since it models the hidden structure of latent signals sources. We attempt to explain why attention mechanisms can achieve great performance in constructing FBNs from the perspectives of cognitive neuroscience and statistical correlations. 

From a cognitive neuroscience perspective, neural activities can be directly measured by BOLD signals. When neurons are excited, the BOLD signals will rise, otherwise the BOLD signals will remain at around the baseline level. Neural activities lead to differences in fMRI signals at each time step. To achieve a good feature representation, the encoder must take into account neural activity, which is the key factor that affects fMRI signal changes. Thus, the attention mechanism will focus on the regions of neural activities.

\begin{figure}
	\centering
	\includegraphics[width=\textwidth]{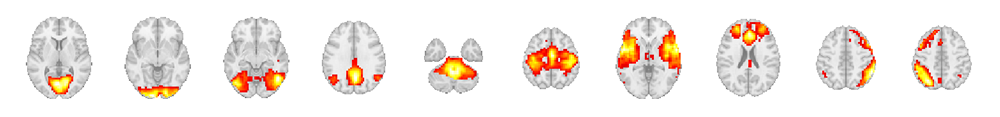}
	\caption{The FBN templates constructed by ICA \cite{3}. The activated voxels are concentrated in the regions nearby themselves. The strong spatial correlation enables the attention mechanism to capture the features of neural activities.}
	\label{fig2}
\end{figure}

From a statistical correlation perspective, activated voxels derived by correlation-based functional connectivity methods are concentrated in the regions near themselves as shown in Figure~\ref{fig2}. The activated regions show strong local correlations. Because locality is an inherent inductive bias for convolution neural networks, using convolution neural networks to model the fMRI signals is a natural choice. 

\subsubsection{Spatial-wise Attention}
The spatial-wise attention (SA) can be represented as:
\begin{equation}
	SA = sigmoid(conv3D(GELU(conv3D(X_{n})))) 
\end{equation}
\begin{equation}
	X_{n+1} = X_{n} \circ SA 
\end{equation}
where $X_{n}$ refers to the output features of the number of convolutional layers, $conv3D$ refers to the 3D convolutional layers, $GELU$ \cite{43} and $sigmoid$ refer to the activation functions. The $sigmoid$ function can map the features to [0, 1]. We can obtain the weighted features map through spatial-wise attention. However, fMRI signals may be from a mixture of two or more functional networks \cite{li2014including,allen2014tracking}. We use two branch networks to discover different FBNs. Specifically, we insert the SA into each branch and then concatenate the output together:
\begin{equation}
	X_{n+1} = concatenate(SA_{1}(X_{n}), SA_{2}(X_{n})) \circ X_{n}
\end{equation}

\subsubsection{Channel-wise Attention}
We weigh the importance of each channel by channel-wise attention (CA) which can be represented as:
\begin{equation}
	CA = sigmoid(\sigma(\sigma(GAP(X_{n}){W_{1}}){W_{2}}){W_{3}})
\end{equation}
where $W$ refers to the fully connected layer, $\sigma$ refers to the activation function GELU and $GAP$ refers to global average pooling.

\subsection{FBN}
\label{2.2}
Each output of the two SA modules is consist of 16 channels of 3D feature maps. Because the sigmoid function maps larger outputs to 1 and smaller outputs to 0, the activated weight map is included in the feature map before it is activated by the sigmoid function. The feature map is defined as $S^{C \times D \times H \times W}$, where $C$ refers to the number of channels, $D$ refers to the depth of the feature map, $H$ refers to the height of the feature map, and $W$ refers to the width of the feature map. Then the $CA^{C \times 1}$ is used to weigh each channel:
\begin{equation}
	S_{weighted} = CA \circ S
\end{equation}
\begin{equation}
	FBN = \sum_{n=1}^{C} S_{weighted}^{n}
\end{equation}
where $S_{weighted}$ refers to the weighted channels by $CA$, $n$ refers to the number of channels. In order to display the FBNs more clearly, we normalize it with Min-Max normalization and square the result to make the activated regions more visible.
\begin{equation}
	FBN = (\frac{FBN - min(FBN)}{max(FBN) - min(FBN)})^2
\end{equation}

\section{Experiments}
\label{3}
\subsection{Dataset and Preprocessing dataset}
\label{3.1}
Attention Deficit Hyperactivity Disorder (ADHD) is a mental health disorder that affects at least 5-10\% of school-age children. The ADHD200\footnote{ADHD200 dataset \url{http://fcon_1000.projects.nitrc.org/indi/adhd200/}} released 904 resting-state fMRI (rsfMRI) and anatomical datasets aggregated across 8 independent imaging sites. The datasets of these 8 sites are released by Kennedy Krieger Institute (KKI), Peking University (PU), New York University Medical Center (NYU), Brown University (BU), Oregon Health \& Science University (OHSU), University of Pittsburgh (UP), Washington University (WU), and NeuroImage (NI) respectively. Each site has different numbers of subjects, as shown in Table~\ref{tab1}. The entire ADHD200 dataset includes 554 typically developing individuals and 350 ADHD patients between the ages of 7 and 12 years. We used preprocessed data publicly available from Preprocessed Connectivity Project \cite{22} for all experiments. The Athena preprocessing pipeline, which is based on AFNI and FSL, was adopted to process the dataset \cite{33,34,35,36,37}. It includes skull stripping, slice time correction, motion correction, detrending, band filtering, normalization and masking. All subjects are registered to the MNI152 $4\times4\times4$ $mm^{3}$ standard template space.

\begin{table}
	\caption{Summary of ADHD200 dataset}\label{tab1}
	\centering
	\begin{tabular}{cccc}
		\toprule
		Imaging Site &  Total Subject & Control Subject & ADHD Subject\\
		\midrule
		KKI &  83 & 61 & 22\\
		PU &  259 & 146 & 113\\
		NYU & 222 & 99 & 123\\
		NI & 48 & 23 & 25\\
		BU & 51 & 27 & 24\\
		OHSU & 79 & 42 &37\\
		UP & 101 & 95 & 6\\
		WU & 61 & 61 & 0\\
		Total Sites & 904 & 554 & 350 \\
		\bottomrule
	\end{tabular}
\end{table}

In accordance with HIPAA guidelines and 1000 Functional Connectomes Project protocols, all datasets are anonymous, without any protected health information.

\subsection{Implementation Details}
\label{3.2}
The experiments were performed on a workstation with 2 NVIDIA GeForce GTX 1080 Ti GPUs. We preprocess the 3D fMRI images with a mask and resample the images to $49 \times 58 \times 47$. Before inputting the data to the attention modules, we do not perform any downsample operations and only use 3D convolutional layers for feature extraction. In our experiments, construction of FBNs can be affected by the number of 3D convolutional layer used prior to the attention mechanism. Specifically, the fewer the convolutional layers are, the closer the FBN construction result is to the linear model. We think this is caused by too many nonlinear transformations associated with the convolutional layers. Therefore, to make the results comparable to the results of other models, we use two convolutional layers, but increase the width of the convolutional layers. Each convolutional layer consists of three convolutional kernels with kernel size 5, 3 and 1 respectively. We also add batch normalization layers (BN) and residual connections after the convolutional kernels. 

When training SCAAE, we use the Adam \cite{39} optimizer with an initial learning rate of 0.001, and we decrease the learning rate with the step decay strategy. The step size is set to 2, and the decay coefficient gamma is set to 0.9. We use mean squared error (MSE) as the loss function. Here $y$ refers to the raw fMRI signals, $\hat{y}$ refers to the reconstructed fMRI signals.
\begin{equation}
	MSE = \frac{1}{N}\sum_{n=1}^{N}(y - \hat{y})^2
\end{equation}

The batch size is set as 12, and epochs is set as 20. We found that the FBNs constructed by the model saved at each epoch were nearly identical. The activation function in each layer is $GELU$.

\subsection{FBNs from SCAAE}
\begin{figure}
	\centering
	\includegraphics[width=\textwidth]{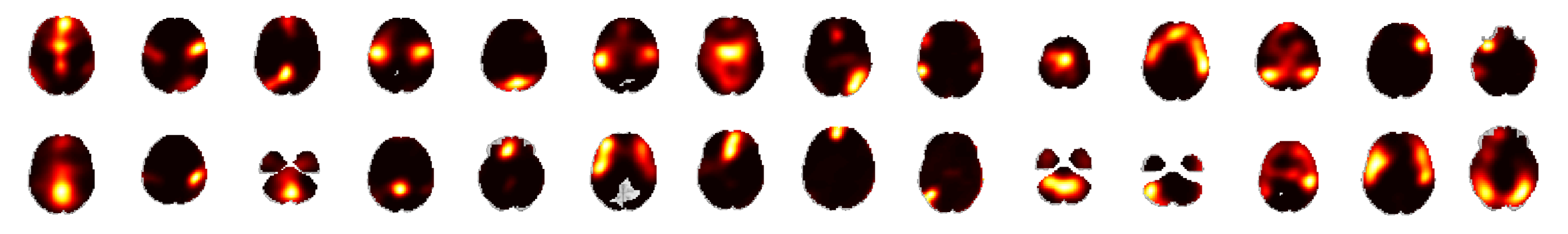}
	\caption{Overview of the FBNs derived from SCAAE. These are the outputs we obtain from the attention modules, and we can observe that the attention mechanism finds the activation regions perfectly.}
	\label{fig3}
\end{figure}

In our experiments, we observe that the brain is constantly active even in the resting state. The FBNs derived from SCAAE are shown in Figure~\ref{fig3}. Since previous work used linear statistical models, the constructed FBNs usually contain a lot of noise and require thresholding with methods such as ICA, SDL and GLM. We use the attention modules to output the FBNs directly. Here, we show the weight map of the attention modules directly without using any thresholding. It is observed that the resting state networks (RSNs) are active even when there are no extra ongoing tasks, which provides evidence supporting the conclusion in \cite{40,41,42,4}. By visual inspection, these FBNs can be well interpreted, and they agree with domain knowledge of functional network atlases in the literature. Also, we find out that the attention mechanism captures the regions of neural activities very well.

\begin{figure}
	\centering
	\includegraphics[width=\textwidth]{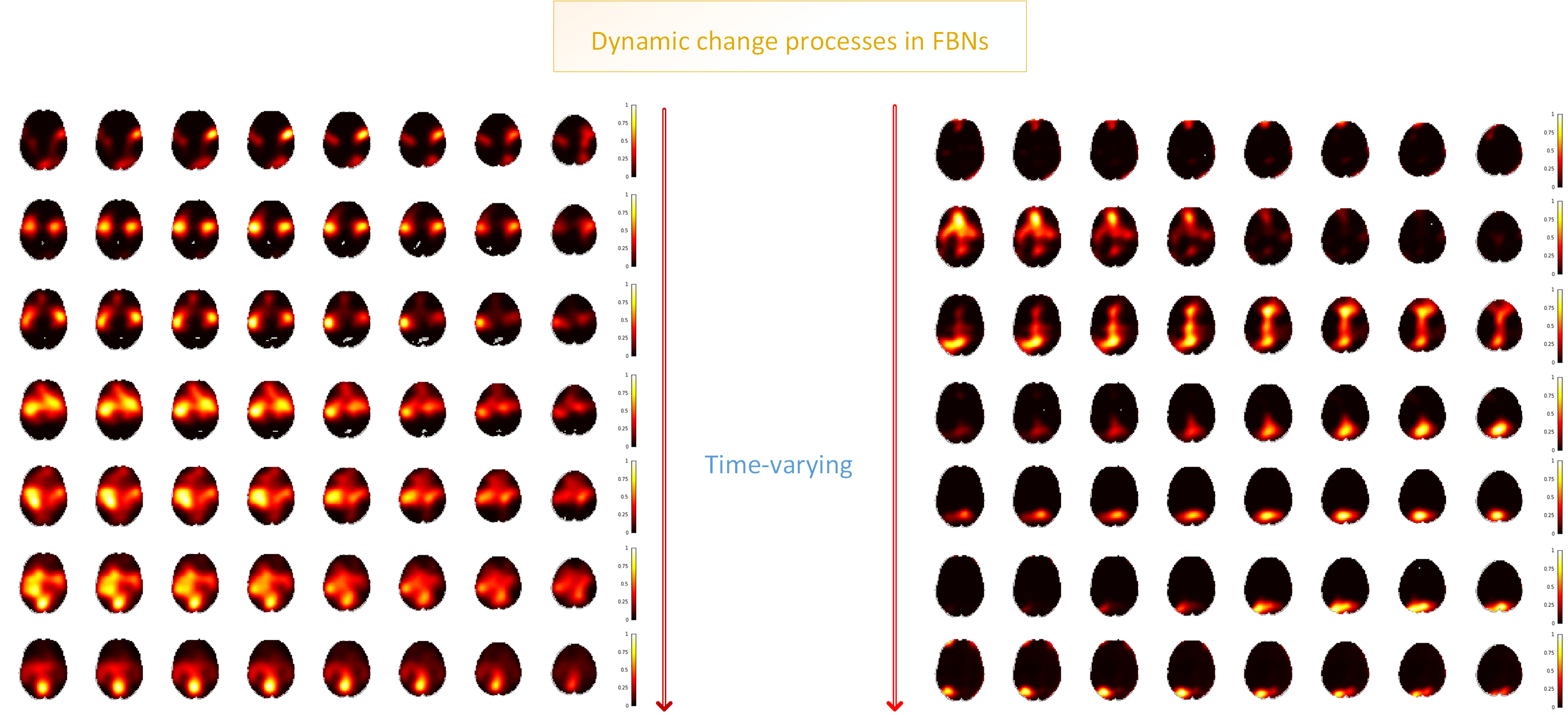}
	\caption{The method can observe changes in FBNs at each time step. These two figures show the state transition process of FBNs and can observe a gradual change process. This is what previous methods cannot provide.}
	\label{fig4}
\end{figure}

We can also observe the transferring process of FBNs in different states. We select two representative transfer processes from individual data and show them in Figure~\ref{fig4}. 

 We measure the similarity of two spatial maps to evaluate the performance of our method. The spatial similarity is defined by the intersection over union rate (IoU) between two FBNs $N^{(1)}$ and $N^{(2)}$, where $n$ refers to the volume size:
\begin{equation}
	IoU(N^{(1)}, N^{(2)}) = \frac{\sum_{i=1}^{n}|N_{i}^{(1)} \cap N_{i}^{(2)}|}{\sum_{i=1}^{n}|N_{i}^{(1)} \cup N_{i}^{(2)}|}
\end{equation}

In addition to subjectively observing the gradual change process of FBNs, we also made a quantitative evaluation of the change of FBNs. We calculated the IoU of the FBN of each individual at each time step versus the previous time step, and averaged all the IoUs to evaluate the change at each time step. The gradual change process of FBNs has been evaluated on 4 subsets (from 4 sites) of the ADHD200 dataset, as shown in Table~\ref{tab2}. The higher the average IoU is, the more it can indicate that the change of the functional networks is gradual. Our results indicates that the change of FBNs is indeed gradual.
	\begin{table}
		\caption{The average IoU of the FBN of each individual at each time step versus the previous time step.}\label{tab2}
		\centering
		\begin{tabular}{ccccc}
			\toprule
			 Site & KKI & PU & NYU & NI\\
			\midrule
			 Average IoU & 0.4071  & 0.4955 & 0.5024	& 0.5103 \\	
			\bottomrule
		\end{tabular}
	\end{table}

Because our method is the first work in the field that can derive the FBNs for each time step, we use the standard templates derived by ICA \cite{3} on a large amount of data as the benchmark. We select the FBNs matching the standard templates from all time steps of an individual and calculate the spatial similarity between them. All time-steps of the individual were then group-analyzed with ICA and SDL to calculate the spatial similarity between them and the standard templates. The similarity metrics $IoU(N_{SCAAE}, N_{template})$, $IoU(N_{ICA}, N_{template})$ and $IoU(N_{SDL}, N_{template})$ are used to evaluate the performance of SCAAE. 

\begin{figure}
	\centering
	\includegraphics[width=0.9\textwidth,height=0.9\textwidth]{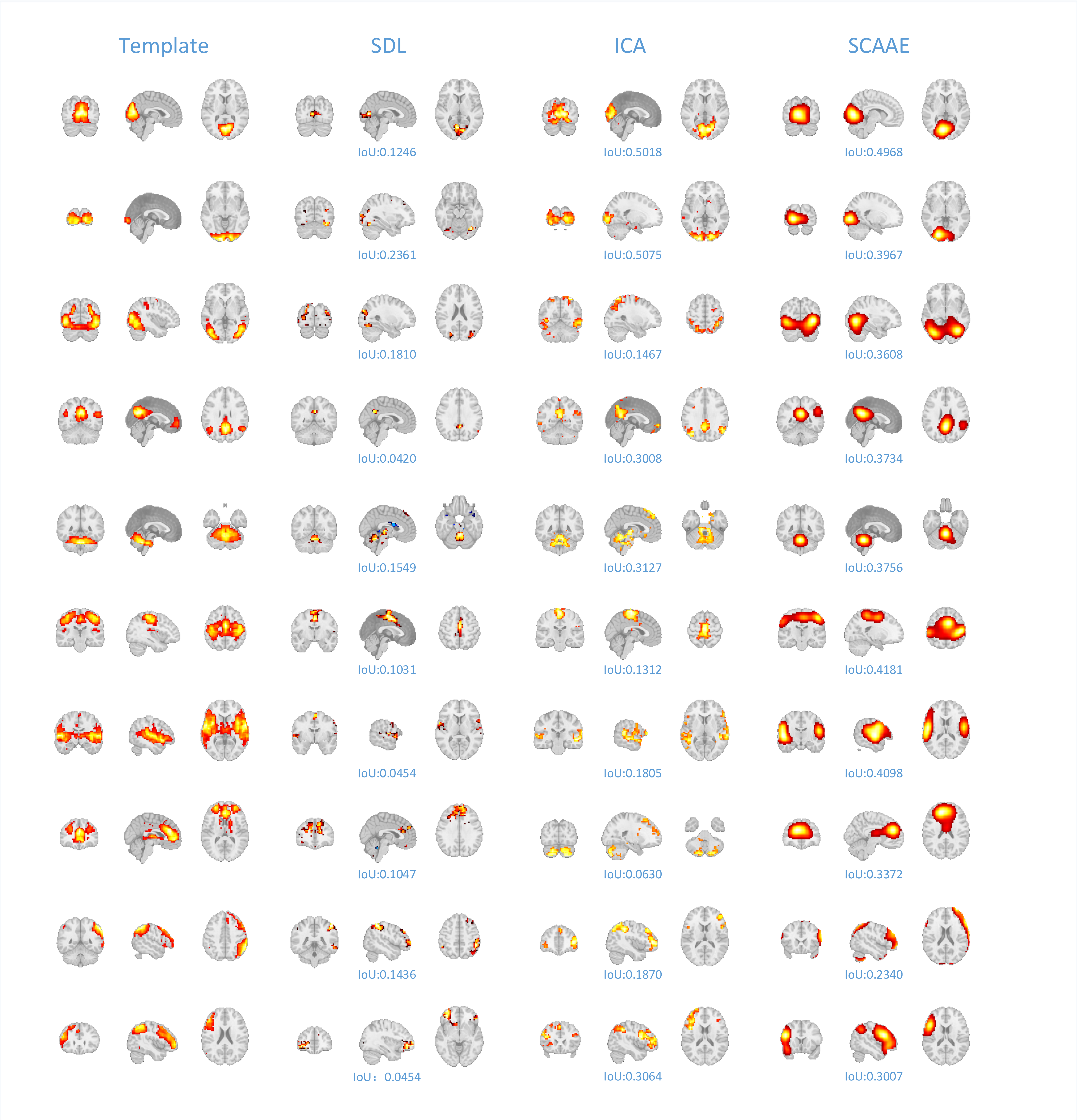}
	\caption{Comparison with SDL and ICA. The IoU refers to the spatial similarity of the corresponding methods and templates. The templates are based on ICA \cite{3}.}
	\label{fig5}
\end{figure}

The results of the evaluation are shown in Figure~\ref{fig5}. The results indicate that SCAAE can identify FBNs very well, suggesting the effectiveness and meaningfulness of our proposed model, even though this comparison is actually negatively unfair to our method, because our method can only see a single fMRI image at a time, while SDL and ICA are able to do group analysis of data from all time steps of a single individual. Nevertheless, our method still achieves excellent performance.
\begin{figure}
	\centering
	\includegraphics[width=0.81\textwidth]{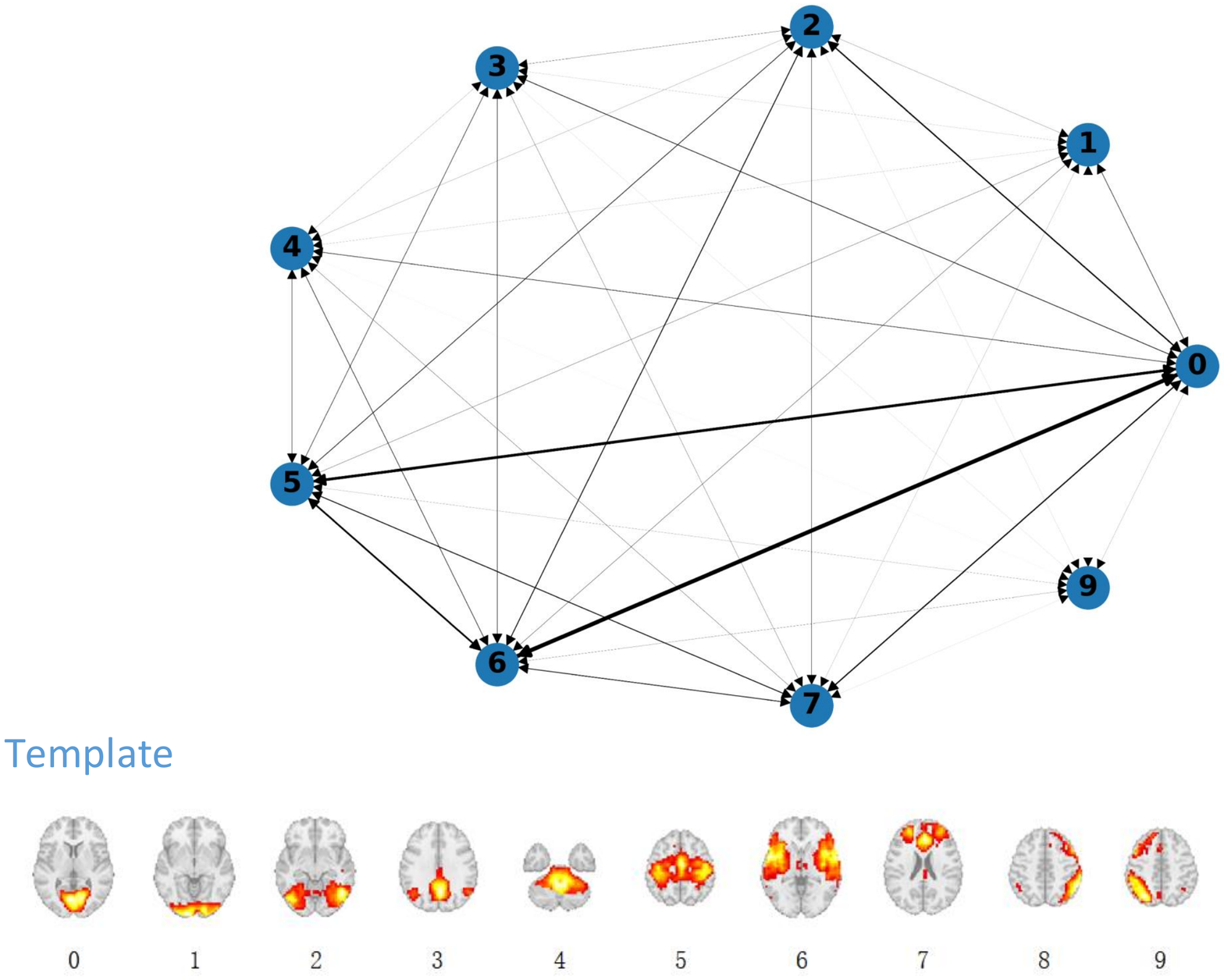}
	\caption{Overview of the functional network state transfer. }
	\label{fig6}
\end{figure}

\subsection{Functional Network State Transfer}
The proposed model SCAAE provides a new perspective to observe the dynamic changes of brain function. We visualize the state transfer of the functional networks. Due to the complexity of human brain, there is no easy way to determine the state of the functional networks. To solve this problem, we use IoU to evaluate the similarity between the functional networks and the templates at each time step, and classify the functional network into the class of its most similar template. The functional networks at each time step can be mapped to one of these ten templates \cite{3}. We filter out the functional networks with IoU less than 0.3. The state transfer diagram is shown in Figure~\ref{fig6}. Thicker lines indicate more frequent transitions between the two states. The number in the figure corresponds to the state of the template. We can observe that the state of template 0 occurs the most and is most closely connected to other states. Template 8 has not appeared yet, which means that it is probably just a theoretical template estimated from a large amount of data, and does not actually exist in individual brain's activities.

\section{Discussion and Conclusion}
\label{4}
In this paper, we proposed a novel method named SCAAE to discover dynamic functional brain networks via visual attention. This work is the earliest study in the field to explore dynamic FBNs with attention mechanism, to the best of our knowledge. Compared with the statistical machine learning models used in previous studies, our method is free from the time-invariant, linear and independent constraints. In addition, since the proposed model only needs to perform forward inference once in the stage of constructing FBNs, our model can construct FBNs faster than the previous methods. The greatest contribution is that we provide a dynamic perspective to observe the changing patterns of FBNs, which is a novel method to understand the human brain.

One limitation of our work is that we have only performed experiments on resting-state fMRI (rsfMRI) but not on task-based fMRI (tfMRI) yet. Although the model can be directly applied to task-based fMRI data, we believe that it is more suitable to develop a new network structure based on the attention mechanism that adapts to the task design requirements of task-based fMRI. In addition, our work is based on autoencoders, and therefore we can also apply the encoder representation as biomarkers to brain disorder characterization, such as ADHD, Alzheimer’s disease, etc. Finally, we believe studying the transition of functional networks states is an important direction for future dynamic FBNs research, which provides a new perspective for understanding the human brain.
		
	{
		\small
		\bibliographystyle{splncs04}
		\bibliography{ref}

	}


	\section{Appendix}

\end{document}